%% file: main.tex
\documentclass[lettersize,journal,twoside]{IEEEtran}

\input{packages}

\graphicspath{ {pictures/} }
\input{abbreviations}

\usepackage{graphics} % for pdf, bitmapped graphics files
\usepackage{amsmath} % assumes amsmath package installed
\makeatletter
\let\NAT@parse\undefined
\makeatother

\usepackage{url}
\usepackage{subcaption} % For subfigures and subcaptions
\usepackage{booktabs}
\usepackage{xcolor}
\usepackage{colortbl}   % enables \rowcolor, \columncolor

\definecolor{TableBlue}{rgb}{0.17,0.49,0.75}
\definecolor{TableWhite}{rgb}{1,1,1}

\usepackage[hidelinks]{hyperref} %hide the borders of the cross-references
\usepackage[nameinlink,capitalise]{cleveref} % cref package

\title{\LARGE \bf
DoGFlow: Self-Supervised LiDAR Scene Flow via Cross-Modal Doppler Guidance
}

\author{Ajinkya Khoche$^{1,2}$, Qingwen Zhang$^{1}$, Yixi Cai$^{1}$, Sina Sharif Mansouri$^{2}$ and Patric Jensfelt$^{1}$% <-this % stops a space

\thanks{$^{1}$KTH Royal Institute of Technology, Stockholm 10044, Sweden. Corresponding author's e-mail: {\tt\small khoche@kth.se}}%
\thanks{$^{2}$Autonomous Transport Solutions Lab, Scania Group, Södertälje, SE-15139, Sweden}
}

\IEEEaftertitletext{\vspace{-1\baselineskip}}
\begin{document}

%%%%%%%%%%%%%%%%%%%%%%%%%%%%%

\maketitle
\thispagestyle{empty}
\pagestyle{empty}

%%%%%%%%%%%%%%%%%%%%%%%%%%%%%
\input{sections/abstract}
\glsresetall

%%%%%%%%%%%%%%%%%%%%%%%%%%%%%
\input{sections/introduction}
\input{sections/related_work}
\input{sections/background}

%%%%%%%%%%%%%%%%%%%%%%%%%%%%%
%\section{Methodology} 
%\label{sec:method}
\input{sections/methodology}
\input{sections/expt_setup}
\input{sections/expt_results}

\input{sections/conclusions}
%%%%%%%%%%%%%%%%%%%%%%%%%%%%%


%%%%%%%%%%%%%%%%%%%%%%%%%%%%%%%%%%%%%%%%%%%%%%%%%%%%%%%%%%%%%%%%%%%%%%%%%%%%%%%%

%\FloatBarrier
%%%%%%%%%%%%%%%%%%%%%%%%%%%%
% Disabled for anonymous submission
\section*{ACKNOWLEDGMENT}
\label{sec:acknowledgment}
We are grateful to Laura Pereira Sánchez for insightful discussions and helpful feedback on the manuscript. This work was supported by the research grant PROSENSE (2020-02963) funded by VINNOVA. The computations were enabled by the supercomputing resource Berzelius provided by National Supercomputer Centre at Linköping University and the Knut and Alice Wallenberg Foundation, Sweden.
%%%%%%%%%%%%%%%%%%%%%%%%%%%%

%\addtolength{\textheight}{-12cm}
\bibliographystyle{IEEEtran}
\bibliography{mybib}

\end{document}

%% file: packages.tex
\usepackage{lineno}
\usepackage{pgfplots}
\pgfplotsset{compat=1.18} 
\usepackage{tikz}
\usetikzlibrary{positioning}
\usetikzlibrary{shapes.geometric}
\usetikzlibrary{shapes.misc}
\usetikzlibrary{external}
\usepackage{paralist}
\usepackage{mathbbol}
\usepackage{amssymb}    
\usepackage{bm}
\usepackage{siunitx}
\usepackage{multirow}
\usepackage{todonotes}
\usepackage{epstopdf}
\usepackage{epsfig}
\usepackage{float}
\usepackage{amsmath} % assumes amsmath package installed
\usepackage{amssymb}  % assumes amsmath package installed
\usepackage{mathtools}
\usepackage[acronym,nomain,nonumberlist]{glossaries}
\usepackage{url}
\usepackage[ruled,vlined,linesnumbered,noresetcount]{algorithm2e}

\newlength\fwidth

\usepackage{pgfplots}
\pgfplotsset{compat=newest}
\usepackage{tikz}
\usetikzlibrary{positioning}
\usetikzlibrary{shapes.geometric}
\usetikzlibrary{shapes.misc}
\usetikzlibrary{external}
\usetikzlibrary{shapes,arrows,patterns}
\usetikzlibrary{arrows.meta}
\usetikzlibrary{plotmarks}
\usetikzlibrary{calc,decorations.markings,math}

\usepackage{placeins} 
\usepackage{tabularx}

\usepackage[ruled,vlined]{algorithm2e}
\SetKwComment{Comment}{// }{}
\SetCommentSty{commentstyle}

\definecolor{TableBlue}{rgb}{0.17,0.49,0.75}
\definecolor{Cerulean}{rgb}{0,0,0.95}
\definecolor{LimeGreen}{rgb}{0.15,0.65,0.15}
\definecolor{RoyalBlue1}{rgb}{0.25,0.41,0.88}
\definecolor{RoyalBlue2}{rgb}{0.18,0.28,0.92}
\definecolor{Rose}{rgb}{1.0, 0.15, 0.21}
\definecolor{Orange}{rgb}{1.0, 0.5, 0.0}
\definecolor{Gray}{gray}{0.6}
\definecolor{Black}{gray}{0.0}
\definecolor{Purple}{rgb}{0.77,0.12,0.64}
\newcommand{\qw}[1]{{\color{Black}#1}}

%% file: abbreviations.tex
\newacronym{ap}{AP}{Average Precision}
\newacronym{av}{AV}{Autonomous Vehicles}
\newacronym{av2}{AV2}{Argoverse2}
\newacronym{aws}{AWS}{Amazon Web Services}
\newacronym{bev}{BEV}{bird's eye view}
\newacronym{cbgs}{CBGS}{Class-balanced grouping and sampling}
\newacronym{ccl}{CCL}{Connected Component Labelling}
\newacronym{cnn}{CNN}{Convolutional Neural Network}
\newacronym{rnn}{RNN}{Recurrent Neural Network}
\newacronym{gru}{GRU}{Gated Recurrent Unit}
\newacronym{cds}{CDS}{Composite Detection Score}
\newacronym{dnn}{DNN}{Deep Neural Network}
\newacronym{epe}{EPE}{End Point Error}
\newacronym{emc}{EMF}{Ego-motion Flow}
\newacronym{ekf}{EKF}{Extended Kalman Filter}
\newacronym{fd}{FD}{Foreground Dynamic}
\newacronym{fs}{FS}{Foreground Static}
\newacronym{bs}{BS}{Background Static}
\newacronym{fsd}{FSD}{Fully Sparse Detector}
\newacronym{fsf}{FSF}{Fully Sparse Fusion}
\newacronym{gps}{GPS}{Global Positioning System}
\newacronym{hd}{HD}{High Definition}
\newacronym{htc}{HTC}{Hybrid Task Cascade}
\newacronym{icp}{ICP}{Iterative Closest Point}
\newacronym{kf}{KF}{Kalman Filter}
\newacronym{lidar}{LiDAR}{Light Detection and Ranging}
\newacronym{map}{mAP}{mean Average Precision}
\newacronym{mav}{MAV}{Micro Aerial Vehicle}
\newacronym{mhe}{MHE}{Moving Horizon Estimation}
\newacronym{mlp}{MLP}{Multi Layer Perceptrons}
\newacronym{mvp}{MVP}{Multimodal Virtual Points}
\newacronym{nds}{NDS}{NuScenes Detection Score}
\newacronym{nmhe}{NMHE}{Nonlinear Moving Horizon Estimation}
\newacronym{nms}{NMS}{Non-Maximum Suppression}
\newacronym{pdf}{PDF}{Probability Density Function}
\newacronym{radar}{RADAR}{Radio Detection and Ranging}
\newacronym{rbgs}{RBGS}{Range-balanced grouping and sampling}
\newacronym{rpn}{RPN}{Region Proposal Networks}
\newacronym{sir}{SIR}{Sparse Instance Recognition}
\newacronym{ssc}{SSC}{Submanifold Sparse Convolutions}
\newacronym{ssf}{SSF}{Sparse Scene Flow}
\newacronym{slam}{SLAM}{Simulataneous Localization and Mapping}
\newacronym{soa}{SoA}{State of the Art}
\newacronym{sota}{SOTA}{state-of-the-art}
\newacronym{sph}{SPH}{sparse prediction head}
\newacronym{vvm}{VVM}{Virtual Voxel Mixer}
\newacronym{vfe}{VFE}{Voxel Feature Encoding}

%% file: sections/abstract.tex
\glsunset{lidar}
\glsunset{radar}
\begin{abstract}
\qw{
%
% First sentence: CONTEXT - why now?
%
Accurate 3D scene flow estimation is critical for autonomous systems to navigate dynamic environments safely, but creating the necessary large-scale, manually annotated datasets remains a significant bottleneck for developing robust perception models.
%
% Second sentence: NEED - why does the reader care?
%
Current self-supervised methods struggle to match the performance of fully supervised approaches, especially in challenging long-range and adverse weather scenarios, while supervised methods are not scalable due to their reliance on expensive human labeling.
%
% Third sentence: TASK - what do we do?
%
We introduce DoGFlow, a novel self-supervised framework that recovers full 3D object motions for LiDAR scene flow estimation without requiring any manual ground truth annotations.
%
% Forth sentence: OBJECT - what does this document do?
%
This paper presents our cross-modal label transfer approach, where DoGFlow computes motion labels directly from 4D radar Doppler measurements and transfers them to the LiDAR domain using dynamic-aware association and ambiguity-resolved propagation.
%
% Sentence 5: FINDINGS - what did we discover?
%
On the challenging MAN TruckScenes dataset, DoGFlow substantially outperforms existing self-supervised methods and improves label efficiency by enabling LiDAR backbones to achieve over 90\% of fully supervised performance with only 10\% of the ground truth data.
For more details, please visit \color{blue}\url{https://ajinkyakhoche.github.io/DogFlow/}\color{black}.
}

\end{abstract}

%% file: sections/introduction.tex
\section{Introduction} 
\label{sec:intro}

%%%%%%%%%%%%%%%%    RESET GLOSSARIES from abstract
\glsunset{lidar}
\glsunset{radar}

Scene flow refers to the dense 3D motion field of points in a dynamic scene, providing crucial priors for perception tasks such as object detection~\cite{lentsch2024union}, multi-object tracking~\cite{pan2024ratrack}, instance segmentation~\cite{chen2025semanticflow}, and motion prediction~\cite{murhij2024ofmpnet}.
Numerous methods have been proposed to improve scene flow estimation, focusing on handling large-scale point clouds, improving accuracy, and reducing runtime~\cite{jund2021scalable,zhang2024deflow,khoche2025ssf,kim2025flow4d}.
% Despite this progress, two core challenges remain. The first is data sparsity, which arises at long ranges or under severe occlusions. These long-tail scenarios are underrepresented in annotated datasets~\cite{khoche2024towards}, making supervised training unreliable or ineffective.
\qw{
Despite this progress, several key challenges remain, especially concerning data sparsity and performance in adverse weather. Data sparsity, for instance, arises at long ranges or under severe occlusions. These long-tail scenarios are underrepresented in annotated datasets~\cite{khoche2024towards}, making supervised training unreliable or ineffective.}
To address this, a variety of self-supervised methods have emerged, which scale better by leveraging raw unlabeled data and, in some cases, even outperform fully supervised models~\cite{vedder2023zeroflow, vedder2024neural}. 
However, these methods typically rely on geometric losses such as Chamfer distance~\cite{li2021neural, zhang2024seflow} or cycle consistency~\cite{mittal2020just, zhang2025himo}, which assume one-to-one correspondences across time. 
In sparse regions, such correspondences may be ambiguous or absent, leading to distorted flow estimates. 
To improve robustness, recent works incorporate smoothness constraints~\cite{luo2021self, jiang20223} or multi-frame optimization~\cite{vedder2024neural}, but at the cost of over-smoothing motion boundaries or increased computation~\cite{hoffmann2025floxels}.

\input{figures/cover_fig}

% A second challenge is adverse weather, such as rain, snow, fog, or dust, as it introduces significant noise into \gls{lidar} point clouds by scattering and absorbing laser pulses, leading to spurious near-field returns, missing data, and high-density low-intensity noise~\cite{Dreissig2023SurveyOL}.
\qw{Adverse weather presents another significant challenge by introducing significant noise into \gls{lidar} point clouds. These noisy conditions, caused by rain, snow, or dust scattering and absorbing laser pulses (as shown in \cref{fig:cover_fig}), lead to spurious near-field returns, missing data, and high-density low-intensity noise~\cite{Dreissig2023SurveyOL}.}
This degradation compromises the assumptions underlying self-supervised losses, resulting in unreliable scene flow estimations. 
While denoising techniques ranging from statistical outlier removal~\cite{wang2022scalable} to deep learning methods~\cite{yu2022lisnownet} have been proposed, their integration into scene flow estimation pipelines remains limited. 
Moreover, such methods may inadvertently remove valid returns or increase computational complexity~\cite{wang2022scalable}.

To address both challenges, another class of self-supervised methods has emerged: cross-modal supervision, which leverages complementary sensor modalities to provide additional constraints during training. 
Among possible modalities, radar is particularly well-suited for outdoor scene flow due to its robustness to adverse weather and its ability to directly measure radial motion through Doppler. 
Radar complements \gls{lidar}, which offers better spatial resolution and dense structure, but lacks direct motion cues.
Although prior work has used Doppler to supervise radar-only scene flow~\cite{ding2022self}, no existing studies have explored 
cross-modal supervision from radar to \gls{lidar}.

In this paper, we extend the idea of cross-modal supervision by investigating whether Doppler measurements from modern 4D millimeter-wave (mmWave) radar can enable direct cross-modal \textit{label transfer} for \gls{lidar} scene flow estimation. 
We recover full 3D velocities by combining Doppler observations with cluster-level rigidity constraints, then propagate these motions to the \gls{lidar} domain as scene flow estimates. 
This enables both real-time inference and offline pseudo-label generation without human annotations.
To the best of our knowledge, we are the first to propose using cross-modal label transfer from radar to \gls{lidar} for scene flow. 
While established benchmarks~\cite{caesar2020nuscenes, jund2021scalable, wilson2argoverse} have been widely used for scene flow, they lack 4D radar data. We instead evaluate on the MAN TruckScenes dataset~\cite{fent2024man}, which uniquely combines long-range \gls{lidar} with 4D radar, enabling rigorous evaluation under long-range and adverse weather conditions.
Furthermore, we show that radar-derived pseudo-labels enable effective self-supervised training of \gls{lidar}-based scene flow networks, improving label efficiency and outperforming prior pseudo-labeling methods~\cite{li2021neural, lin2024icp}. The code will be open-sourced at \color{blue}\href{https://github.com/ajinkyakhoche/DoGFlow}{https://github.com/ajinkyakhoche/DoGFlow}\color{black}~upon acceptance.

To summarize, our contributions are:
\begin{itemize}
    \item We introduce DoGFlow, a cross-modal scene flow framework that recovers full 3D velocities from Doppler-aware clustering and propagates them to \gls{lidar} via range-adaptive dynamic association.
    
    \item We demonstrate that the resulting pseudo-labels enable self-supervised training of existing \gls{lidar}-based scene flow networks, achieving strong performance even with limited ground truth.

    \item We conduct extensive evaluations on the MAN TruckScenes dataset, showing that DoGFlow achieves state-of-the-art scene flow performance at long ranges and maintains robustness under adverse weather conditions such as rain and snow (see~\Cref{fig:cover_fig}).
\end{itemize}

%% file: figures/cover_fig.tex
\begin{figure}[th]
\centering
\includegraphics[trim=0 0 0 0, clip, width=0.97\linewidth]{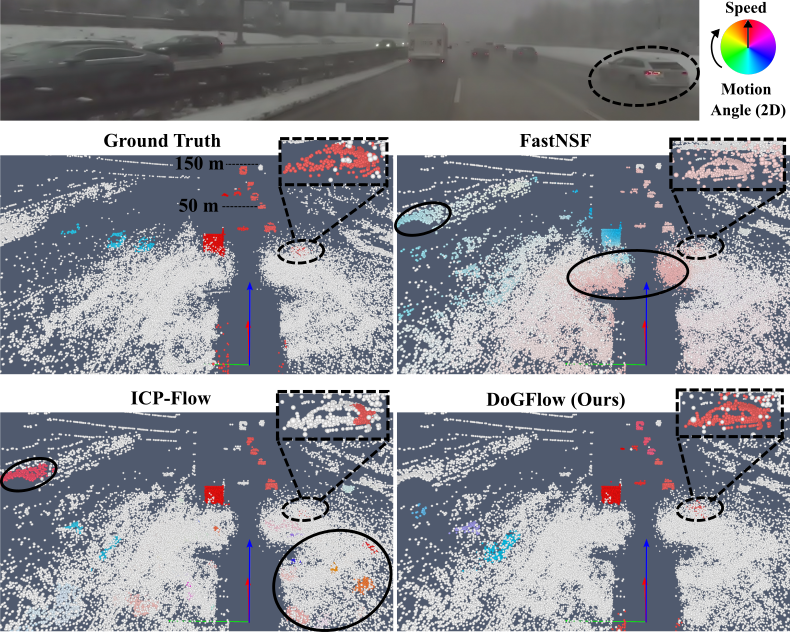}
\caption{Qualitative scene flow comparison on a real-world snowy scene from the MAN TruckScenes dataset. Methods relying on geometric point matching produce noisy (solid ellipse) or inaccurate (dashed ellipse) estimation or miss completely at long-range. In contrast, DoGFlow combines noise-resistant clustering with radar-guided motion cues to deliver accurate scene flow under adverse weather and sparse data conditions.}
\label{fig:cover_fig}
\vspace{-1.2em}
\end{figure}

%% file: sections/related_work.tex
\section{Related Work}
\label{sec:related_work}
\subsection{Self-supervised Scene Flow}
An early work by Mittal et. al.~\cite{mittal2020just} proposed using nearest neighbor matching and cycle consistency losses as a self-supervision signal for scene flow on large-scale autonomous driving datasets. 
Building on the advancements in neural rendering approaches, Li et. al.~\cite{li2021neural} formulated scene flow as a runtime optimization problem, 
% and proved that a simple neural network can be used to solve it. 
leverging a simple neural network as a solution, allowing for flexible adaptation to new environments without retraining.
To enhance computational efficiency, subsequent works introduced techniques such as distance transform matching~\cite{li2023fast}, reducing the reliance on computationally intensive operations like Chamfer distance calculations.
% Recent methods have proposed using multiple timestamp information~\cite{hoffmann2025floxels} or posing scene flow as a euler field~\cite{vedder2024neural}.
Recent methods~\cite{hoffmann2025floxels, vedder2024neural} aim to exploit temporal information, optimizing the neural representation against multi-observation reconstruction objectives.

% Alternative approaches incorporate local rigidity assumptions~\cite{li2022rigidflow} and geometric constraints with handcrafted initialization schemes~\cite{lin2024icp}, to align point correspondences between frames. These methods generate pseudo-labels that can supervise real-time compatible models, bridging the gap between traditional optimization techniques and supervised learning approaches.
Alternative approaches formulate scene flow as a piecewise rigid motion estimation task, segmenting the point cloud into locally rigid regions and aligning each via rigid registration~\cite{li2022rigidflow}. ICP-Flow~\cite{lin2024icp} applies classical Iterative Closest Point (ICP) with a histogram-based motion prior to robustly align \gls{lidar} clusters across frames. These methods generate pseudo-labels that can supervise real-time compatible models, bridging the gap between traditional optimization techniques and supervised learning approaches.
Dynamic object awareness~\cite{duberg2024dufomap} has been another area of exploration. SeFlow~\cite{zhang2024seflow} leverages this by focusing time-intensive clustering only on dynamic points, and further performing cluster matching between sequential scans, estimating maximal cluster motion to serve as pseudo-labels.
% Despite these advancements, most self-supervised methods fail to converge for sparse point sets or under adverse weather conditions, where sensing noise may lead to loss of geometric information, and thereby the self-supervision signal.
Despite these advancements, many self-supervised methods struggle in scenarios characterized by sparse point distributions or degraded sensor data, where the assumptions underpinning their loss functions—such as reliable point correspondences—no longer hold, leading to suboptimal performance.

% \subsection{Radar-only Scene Flow}
% Current automotive radars produce much sparser point clouds than \gls{lidar}, making direct scene flow estimation particularly challenging. RaFlow~\cite{ding2022self} adopts a two-stage pipeline that first estimates coarse per-point flow using relative Doppler velocities, and then applies rigid alignment via the Kabsch algorithm to compensate for ego-motion. It employs Chamfer distance, smoothness constraints, and a motion-consistency loss for self-supervised training. 
% RadarMOSEVE~\cite{pang2024radarmoseve} introduces a spatial-temporal transformer network that leverages both self-attention and cross-attention to extract structured motion features from radar data. It is trained for radar-only moving object segmentation and ego-velocity estimation using supervised losses.
% TARS~\cite{wu2025tars} further incorporates traffic-aware context by leveraging pre-trained object detectors to inform radar scene flow estimation, improving semantic consistency. 
% Despite these efforts, radar-only methods often struggle due to low spatial resolution, sparse detections, and lack of rich supervision signals.

\subsection{Cross-Modal Supervision for Scene Flow}
Several works explore supervising scene flow using complementary sensor modalities. Luo et al.~\cite{luo2021self} propose to use camera based optical flow alongside self-supervised losses to guide \gls{lidar} scene flow. 
EmerNeRF~\cite{yang2023emernerf} decomposes scenes into static and dynamic components and uses emergent scene flow fields to aggregate temporal features, all without explicit motion supervision.
Long et al.~\cite{long2021full} derive a closed-form solution for recovering full 3D radar velocity vectors by combining Doppler-based depth with camera optical flow, and demonstrate its utility for evaluating velocity accuracy, radar point accumulation, and object pose estimation. Ding et. al.~\cite{ding2023hidden} further extends this idea by integrating multiple weak signals, including odometry, camera flow, and \gls{lidar} scene flow, to generate high-quality pseudo-labels for radar supervision. 
While optical flow provides rich 2D motion cues, it mainly captures motion within the image plane and struggles to supervise motion along the depth axis. This limitation becomes especially pronounced at long distances, where objects appear small and motion estimation becomes less reliable.

In contrast to prior work, our method derives full 3D velocity vectors for radar points without relying on camera optical flow~\cite{long2021full} or multi-sensor supervision signals during training~\cite{ding2023hidden}. Instead, we leverage the increased density of modern 4D radar to cluster dynamic radar points and estimate cluster-level rigid motions geometrically. These motions are then transferred to \gls{lidar} domain as scene flow labels, a form of cross-modal label transfer that does not require model training.
% This allows us to generate full-scene velocity estimates, which we transfer to \gls{lidar} for scene flow estimation, a direction not previously explored. Moreover, our method is lightweight and modular: 
Our approach is lightweight and modular. It can be deployed at runtime to estimate scene flow directly, or used offline to generate pseudo-labels for training \gls{lidar}-based scene flow networks without requiring radar at inference time.

%% file: sections/background.tex
\section{Background}
\label{sec:background}
\subsection{Problem Statement}
The task of scene flow estimation aims to recover a point-wise 3D displacement field \( \hat{\mathcal{F}}_t \in \mathbb{R}^{N \times 3} \) for a given point cloud \( \mathcal{P}_t \) such that the warped point cloud \( \hat{\mathcal{P}}_{t+1} = \mathcal{P}_t + \hat{\mathcal{F}}_t \) approximates the observed point cloud \( \mathcal{P}_{t+1} \). 
The predicted flow can be decomposed as:
\begin{equation}
    \hat{\mathcal{F}}_{t} = \mathcal{F}_{ego} + \Delta \hat{\mathcal{F}}_{t},
\end{equation}
where \( \mathcal{F}_{ego} \) is the ego-vehicle motion, which is assumed known through GPS or a localization pipeline~\cite{vizzo2023kiss}. 
The non-ego flow, \( \Delta \hat{\mathcal{F}}_t \), captures the motion of other agents, and is estimated by a model \( \mathcal{H}_\theta \) (typically a deep neural network) operating on time-sequential scans:
\begin{equation}
    \Delta \hat{\mathcal{F}}_{t} = \mathcal{H}_\theta (\mathcal{P}_t + \mathcal{F}_{ego}, \mathcal{P}_{t+1}).
\end{equation}

The model \( \mathcal{H}_\theta \) can be trained under several distinct learning paradigms. In a supervised setting, the network is trained using ground-truth flow \( \Delta \mathcal{F}^*_t \), typically derived from bounding box tracks. In contrast, self-supervised approaches optimize the network without any flow labels, often by minimizing geometric alignment losses between predicted and target point clouds. 
A common formulation is the chamfer distance loss~\cite{li2021neural, luo2021self, zhang2024seflow}, which quantifies the similarity between two point clouds by measuring the average closest-point distance between them.
Additional losses, such as cycle consistency~\cite{mittal2020just} encourage the predicted forward and backward flows to be mutually consistent, are often used to regularize learning.
An alternative self-supervised strategy is to estimate pseudo scene flow labels $\Delta \mathcal{F}^{\text{pseudo}}_{t}$ through geometric~\cite{lin2024icp} or rigid-motion priors~\cite{li2022rigidflow}, and use them to train the backbone via supervised losses. 
% Our work follows this route.

\qw{To address the core challenges of adverse weather and data sparsity, which compromise the reliability of purely geometric approaches, we turn to complementary modalities by leveraging the direct velocity measurements provided by 4D radar.}
% Given the inherent limitations of geometric losses, especially in sparse or noisy conditions, our work explores an alternative supervisory signal for LiDAR point cloud scene flow estimation: velocity measurements from 4D radar.

\subsection{Radar Multipath Effects and Doppler Anomalies}
\label{subsec:radar_multipath}
\qw{While radar measurements provide a powerful supervisory signal, they are susceptible to physical artifacts that must be addressed. For instance,}
4D mmWave radars rely on range and Doppler Fast Fourier Transforms (FFTs), followed by threshold-based peak detection algorithms~\cite{wang2021cfar}, which can lead to spurious detections that appear in the radar point. 
Mitigating such anomalies is therefore critical for reliable velocity estimation and downstream scene flow tasks. In particular, mmWave radar systems are susceptible to multipath reflections from surfaces such as roads, guardrails, and nearby vehicles. These reflections occur when the radar signal reaches the receiver via multiple propagation paths—e.g., directly from the object and indirectly via ground or structural surfaces~\cite{levy2023mcrb}. When both the direct and multipath signals overlap in time and frequency, they can produce interference patterns that distort the estimated range, angle, or Doppler velocity of a target. This superposition can result in physically implausible measurements, such as ghost targets or contradictory velocity estimates among points belonging to the same physical object~\cite{kopp2021fast}, as shown in~\Cref{fig:multipath_illustration}. 
\qw{These artifacts highlight a core challenge: since individual radar measurements can be unreliable, a robust method must aggregate information from multiple points to recover a consistent motion. The proposed approach, DoGFlow, is designed precisely to tackle this challenge.}
\input{figures/multipath_illustration}

%% file: figures/multipath_illustration.tex
\begin{figure}[ht]
\centering
\includegraphics[trim=0 0 0 0, clip, width=0.95\linewidth]{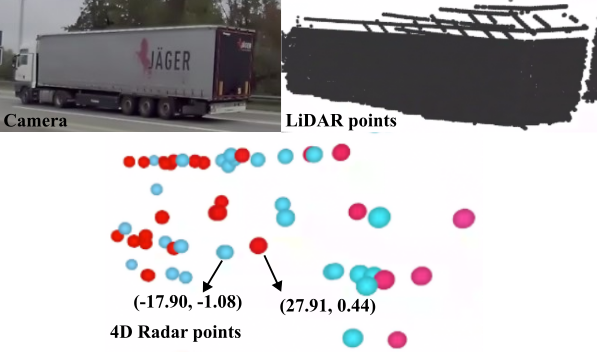}
\caption{Radar Doppler anomalies for a rigid object observed by the rear-left radar. Points on the same truck 
% (outlined in image) 
exhibit conflicting Doppler velocities despite being part of the same physical structure. 
% Point color encodes Doppler direction and magnitude.
Red and blue respectively denote positive and negative doppler signals in radar frame.
}
% A synchronized image from the rear-facing camera shows the corresponding truck in the physical scene.}
\label{fig:multipath_illustration}
\vspace{-1.2em}
\end{figure}

%% file: sections/methodology.tex
\section{Method}
\label{sec:method}
\input{figures/schematic_fig}
\input{algorithms/algorithm_1}
\qw{DoGFlow achieves cross-modal guidance for LiDAR scene flow through a two-stage process, as illustrated in \Cref{fig:schematic_fig}.
The first stage tackles the challenge of \emph{noisy and ambiguous radar data}, estimating robust, cluster-level motion vectors directly from Doppler cues (\Cref{subsec:radar_vel_estimation}). The second stage then addresses the \emph{cross-modal transfer problem}, propagating these high-quality radar-derived motions to the LiDAR domain via noise-aware clustering, dynamic classification, and ambiguity-resolved label propagation (\Cref{subsec:radar_to_lidar_label_prop}).
}

\subsection{Radar Velocity Estimation}
\label{subsec:radar_vel_estimation}
4D radars provide robust sensing capabilities under adverse weather conditions, owing to their longer wavelengths that penetrate atmospheric particles such as rain, fog, and dust. 
Each radar point $r \in \mathcal{P}_{\text{radar}}$ has a position $\mathbf{p}_r \in \mathbb{R}^3$ and a measured \emph{radial} velocity $v_{\text{meas}} \in \mathbb{R}$, i.e., the projection of the true velocity vector onto the line-of-sight unit vector $\mathbf{u}=\mathbf{p}_r/\lVert\mathbf{p}_r\rVert \in \mathbb{R}^3$.
To estimate the true motion of objects, we compensate for the ego-vehicle's movement. 
For each radar sensor $S_i$, we transform the ego-motion vector $\mathbf{v}_{\text{ego}} \in \mathbb{R}^3$ from the vehicle frame to the radar frame using the extrinsic rotation matrix ${}^{S_i}\mathbf{R}_{\text{ego}} \in \mathbb{R}^{3\times 3}$, where \({}^B\mathbf{R}_A\) denotes the rotation matrix that transforms a vector from frame \(A\) to frame \(B\).  
The ego-motion-compensated radial velocity $v_{\text{comp}}\in \mathbb{R}$ is then given by:
\begin{align}
    v_{\text{comp}} 
    = v_{\text{meas}} + \mathbf{u}^\top {}^{S_i}\mathbf{R}_{\text{ego}} \, \mathbf{v}_{\text{ego}}.
\label{eq:radar_ego_motion_compensation}
\end{align}
Thereafter we approximate the set of dynamic radar points as:
\begin{align}
    \mathcal{P}^{\text{dyn}}_{\text{radar}} 
    = \left\{ r^j \in \mathcal{P}_{\text{radar}} \;\middle|\; |v_{\text{comp}}^j| > \delta_{\text{dyn}} \right\},
\end{align}
where $\delta_{\text{dyn}}$ is a fixed motion threshold and $j$ indexes a radar point in the scan. We then cluster only these dynamic points to separate individual moving objects.
4D radar data exhibits lower spatial resolution than \gls{lidar}, making it challenging to distinguish closely situated objects, such as vehicles traveling side by side on a highway. Traditional clustering methods~\cite{mcinnes2017hdbscan} often struggle under these conditions due to the sparsity and noise inherent in radar data. 
Inspired by the approach in~\cite{fan2022fully}, we employ Connected Components Labeling (CCL) on a graph constructed from the dynamic radar points $\mathcal{P}^{\text{dyn}}_{\text{radar}}$. 
In this graph \(G=(V, E)\), each node \( j \in V \) represents a dynamic radar point with position \( \mathbf{p}^j_r \in \mathbb{R}^3 \) and and ego-motion-compensated velocity vector $\mathbf{v}_{\text{comp}}^j = v_{\text{comp}}^j \, \mathbf{u}^j \in \mathbb{R}^3$.
The edges \(E\) are established based on proximity in both spatial and Doppler velocity domains, as: 
\begin{align}
    \lVert \mathbf{p}^j_r - \mathbf{p}^m_r \rVert < \delta_{\text{spatial}}
    \quad \text{and} \quad
    \lVert \mathbf{v}_{\text{comp}}^j - \mathbf{v}_{\text{comp}}^m \rVert < \delta_{\text{velocity}}
\label{eq:ccl}
\end{align}
The connected components in this sparse adjacency graph \(G\) correspond to clusters of radar points $C_R$ likely belonging to the same moving object. This step scales as \( \mathcal{O}(N^2) \) in the number of dynamic points \( N \), due to the need to evaluate all pairwise connections. However, the sparsity of dynamic radar returns ensure tractable runtimes in our setting.
% In this graph, each node represents a dynamic radar point, and edges are established based on proximity in both spatial and Doppler velocity domains. The connected components in this sparse adjacency graph correspond to clusters of radar points likely belonging to the same moving object. While CCL is typically computationally expensive, the sparsity of dynamic radar points ensures tractable runtimes in our setting.

% Following clustering, 
For each dynamic radar cluster $k$, we estimate a full 3D velocity vector \( \mathbf{v}^k_{\text{full}} \in \mathbb{R}^3 \) (or $\mathbf{v}^k$ for brevity), under the assumption that all points within a cluster belong to a rigidly moving object. While 4D radar provides only radial velocity measurements per point, we exploit the diversity of viewpoints within a cluster to recover full motion. 
For each point $j$ in a cluster $k$, the relationship between the full velocity \( \mathbf{v}^k \) and the ego-motion compensated radial velocity \( v^j_{\text{comp}} \), defined in~\Cref{eq:radar_ego_motion_compensation}, is given as:
% \vspace{-5pt}
\begin{align}
    \underbrace{v^j_{\text{comp}}}_{\mathbf{b}} 
    &= \underbrace{(\mathbf{u}^j)^\top {}^{S_i}\mathbf{R}_{\text{ego}}}_{\mathbf{A}} 
       \mathbf{v}^k 
    \label{eq:vobj_to_vcomp}
\end{align}
% \vspace{-5pt}

This allows us to construct a linear system of the form \( \mathbf{A} \mathbf{v}^k = \mathbf{b} \). 
Depending on the number and distribution of points in each cluster, the resulting linear system may be over or under-constrained. We solve for the estimated full velocity \( \Hat{\mathbf{v}}^k \) using a least-squares solver with physical velocity bounds~\cite{branch1999subspace}, that accommodates both regimes and promotes stable estimates under noisy conditions. The resulting estimate \( \Hat{\mathbf{v}}^k \) is assigned to all radar points within the cluster $k$ and the process is repeated for all clusters.

\subsection{Radar to \gls{lidar} Label Propagation} \label{subsec:radar_to_lidar_label_prop}
To propagate motion labels from radar to \gls{lidar}, we employ a multi-step strategy, detailed in~\Cref{alg:label_propagation}. As a preprocessing step, we perform ground removal~\cite{steinke2023groundgrid} and compute per-point nearest-neighbor associations from each \gls{lidar} point $\mathbf{p} \in \mathcal{P}_t$ to the set of dynamic radar points $\mathcal{P}^{\text{dyn}}_{\text{radar}}$, using a range-adaptive threshold $\delta_{\text{adaptive}}(\mathbf{p})$. Points with no valid radar association are considered noisy and handled later.
We then partition the remaining \gls{lidar} points into high- and low-intensity subsets using a fixed intensity threshold $\delta_{\text{intensity}}$. The high-intensity points, along with any low-intensity points that are successfully associated with a radar point, are clustered using HDBSCAN~\cite{mcinnes2017hdbscan}. This ensures that noisy, dense regions from adverse weather are excluded from clustering, preventing erroneous object merging. The previously filtered low-intensity points are then reintroduced by assigning them to the nearest existing cluster within a distance threshold $\delta_{\text{neighbor}}$, striking a balance between robustness and completeness.

To determine dynamic clusters, we query each \gls{lidar} cluster for associated radar points and apply majority voting: 
% over the radar-based dynamic classification labels: 
if more than half of them belong to $\mathcal{P}^{\text{dyn}}_{\text{radar}}$, the entire \gls{lidar} cluster is labeled as dynamic.
For clusters deemed dynamic, we transfer the radar cluster velocities \( \Hat{\mathbf{v}}^k \) estimated in~\Cref{subsec:radar_vel_estimation}. In cases where multiple radar clusters are associated to a \gls{lidar} cluster (as discussed in~\Cref{subsec:radar_multipath}), we forward-project the \gls{lidar} cluster to time $t+1$ using each candidate velocity and select the best fit via Chamfer distance against $\mathcal{P}_{t+1}$. The final per-cluster velocity is then used to compute the scene flow $\Delta\hat{\mathcal{F}}_t$ as shown in the algorithm.

%% file: figures/schematic_fig.tex
\begin{figure*}[ht]
\centering
\includegraphics[trim=0 0 0 0, clip, width=\linewidth]{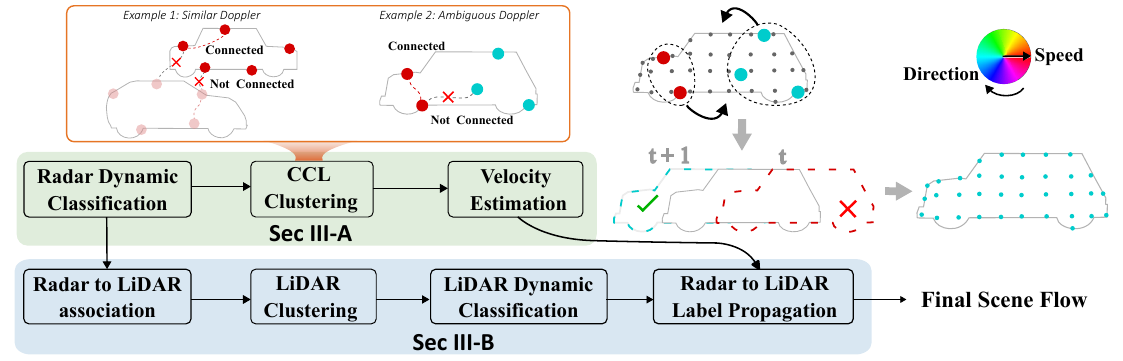}
\caption{Overview of the DoGFlow pipeline. Radar points (larger circles) are first classified as dynamic using ego-motion-compensated Doppler velocity, and clustered via CCL. Cluster-level velocities are estimated using per-cluster linear constraints. Meanwhile, \gls{lidar} points (smaller circles) undergo denoising and clustering. Dynamic labels and velocities are propagated from radar to LiDAR clusters via nearest-neighbor association and majority voting. In cases of velocity ambiguity, candidate motions are resolved via forward projection and Chamfer distance minimization. The resulting radar cluster-level motion is assigned to all \gls{lidar} points in the cluster, yielding dense scene flow estimation.}
\label{fig:schematic_fig}
% \vspace{-1em}
\end{figure*}

%% file: algorithms/algorithm_1.tex
\SetKwInput{KwInput}{Input}                
\SetKwInput{KwOutput}{Output}
\begingroup
\setlength{\textfloatsep}{2pt}
\begin{algorithm}[ht]
\caption{Radar to LiDAR Label Propagation}
\label{alg:label_propagation}

\KwInput{LiDAR scans $\mathcal{P}_t, \mathcal{P}_{t+1}$; Radar clusters $\{C_R^k\}$ with velocities $\{\mathbf{v}^k\}$; Dynamic radar points $\mathcal{P}^{\text{dyn}}_{\text{radar}}$}
\KwOutput{Dense scene flow field $\Delta\hat{\mathcal{F}}_{t}$}

\BlankLine
\Comment{1. LiDAR Preprocessing}
% \State \small{\textcolor{gray}{// 1. LiDAR Preprocessing}}
$\mathcal{P}'_t \gets \mathrm{GroundRemoval}(\mathcal{P}_t)$\;
$\mathcal{P}_{\text{high}} \gets \{\mathbf{p} \in \mathcal{P}'_t \mid \mathrm{intensity}(\mathbf{p}) \ge \delta_{\text{intensity}}\}$\;
$\mathcal{P}_{\text{low}} \gets \mathcal{P}'_t \setminus \mathcal{P}_{\text{high}}$\;

\BlankLine
\Comment{2. Radar-LiDAR Association and Clustering}
% \State \small{\textcolor{gray}{// 2. Radar-LiDAR Association and Clustering}}
$\mathcal{M}(\mathbf{p}) \gets \arg\min_{\mathbf{p}_r \in \mathcal{P}^{\text{dyn}}_{\text{radar}}} \lVert \mathbf{p} - \mathbf{p}_r\rVert$\;
$\mathcal{P}_{\text{assoc}} \gets \{\mathbf{p} \in \mathcal{P}_{\text{high}} \mid \lVert \mathbf{p} - \mathcal{M}(\mathbf{p})\rVert < \delta_{\text{adaptive}}(\mathbf{p})\}$\;
$\{C_L^c\}_{\text{init}} \gets \mathrm{HDBSCAN}(\mathcal{P}_{\text{assoc}})$\;

\ForEach{$\mathbf{p}_{\text{low}} \in \mathcal{P}_{\text{low}}$}{
  $\mathbf{p}' \gets \arg\min_{\mathbf{q} \in \mathcal{P}_{\text{assoc}}} \lVert \mathbf{p}_{\text{low}} - \mathbf{q}\rVert$\;
  \If{$\lVert \mathbf{p}_{\text{low}} - \mathbf{p}'\rVert < \delta_{\text{neighbor}}$}{
    $C_L(\mathbf{p}') \gets C_L(\mathbf{p}') \cup \{\mathbf{p}_{\text{low}}\}$\;
  }
}

\BlankLine
% \State \small{\textcolor{gray}{// 3. Label Propagation}}
\Comment{3. Label Propagation}
Initialize $\Delta\hat{\mathcal{F}}_{t}$ with zeros\;
\ForEach{$C_L^c$}{
  $\{\mathbf{r}^j\} \gets \mathrm{GetAssociatedRadarPoints}(C_L^c)$\;
  \If{$\frac{1}{|\{\mathbf{r}^j\}|} \sum_j \mathbb{I}(\mathbf{r}^j \in \mathcal{P}^{\text{dyn}}_{\text{radar}}) > 0.5$}{
    $\{\mathbf{v}^k\} \gets \mathrm{GetUniqueVelocities}(\{\mathbf{r}^j\})$\;
    \If{$|\{\mathbf{v}^k\}| > 1$}{
      $\mathbf{v}^c_{\text{best}} \gets \arg\min_{\mathbf{v}^k} \; \mathcal{D}_{\text{Chamfer}}\big(C_L^c + \mathbf{v}^k \cdot \Delta t,\; \mathcal{P}_{t+1}\big)$\;
    }
    \Else{
      $\mathbf{v}^c_{\text{best}} \gets \mathrm{first}(\{\mathbf{v}^k\})$\;
    }
    $\Delta\hat{\mathcal{F}}_{t}[C_L^c] \gets \mathbf{v}^c_{\text{best}} \cdot \Delta t$\;
  }
}
\Return{$\Delta\hat{\mathcal{F}}_{t}$}\;
\end{algorithm}
\endgroup

%% file: sections/expt_setup.tex
\section{Experimental Setup}
\label{sec:expt_setup}
\input{tables/table_1}

\subsection{Dataset}
Experiments are conducted on the large-scale autonomous driving dataset, MAN TruckScenes.
Unlike prevailing benchmarks~\cite{caesar2020nuscenes, jund2021scalable, wilson2argoverse}, which lack Doppler-capable radar or offer limited radar resolution, TruckScenes is the first dataset to provide 360° coverage using multiple long-range \glspl{lidar} and modern 4D mmWave radars.
This makes it particularly well-suited for motion estimation in long-range and adverse weather conditions.
The dataset encompasses 747 scenes, each lasting approximately 20 seconds at 10Hz, captured under diverse environmental conditions, including various weather scenarios such as rain and snow, as well as different times of day and lighting situations. The training set consists of 101,902 frames from around 500 scenes.
The annotations in the form of 3D bounding boxes cover every fifth frame.

\subsection{Metrics}
% We employ the following metrics to evaluate scene flow performance: 
We evaluate scene flow performance using the following metrics, all computed over a $409.6\,\si{m} \times 409.6\,\si{m}$ grid centered at the ego-vehicle:

\noindent\textbf{Range-wise EPE}~\cite{khoche2025ssf}, which bins points by their radial distance from the ego-vehicle and computes EPE within each bin. This stratified analysis highlights how flow accuracy is impacted with range—a critical factor for anticipatory perception in autonomous driving. 

\noindent\textbf{Dynamic IoU} measures the Jaccard index~\cite{everingham2015pascal} between predicted and ground-truth dynamic point classifications. A point is labeled as dynamic if its flow magnitude exceeds $0.05\,\si{m}$ over a single frame interval, followed~\cite{wilson2argoverse}. 
Similar to Range-wise EPE, we compute Dynamic IoU across radial bins to evaluate classification consistency at increasing distances. 

\noindent\textbf{Three-way EPE}~\cite{Chodosh_2024_WACV} partitions the point cloud into Foreground Dynamic (FD), Foreground Static (FS), and Background Static (BS) regions. The \gls{epe}, defined as the $L_2$ norm between predicted flow and ground truth, is computed separately for each region and then averaged to yield the final score. 
% Lastly, we also report the Bucket-Normalized \gls{epe}~\cite{khatri2024can}, which organizes points into two-dimensional buckets based on object class and motion speed. The EPE is normalized within each bucket by the average speed of that bucket, allowing for fair comparison across fast- and slow-moving object categories. 

\subsection{Baselines}
We evaluate DoGFlow against self-supervised learning based scene flow methods. NSFP~\cite{li2021neural} is a pioneering method that uses a coordinate-based neural network to optimize scene flow at runtime. FastNSF~\cite{li2023fast} improves the runtime of NSFP using a distance transform-based loss for efficient correspondence matching. ICP-Flow~\cite{lin2024icp} is a learning-free approach that uses histogram-based initialization for \gls{icp} to estimate local rigid transformations. Additionally, we compare to SeFlow, 
% (described in~\Cref{subsec:dy_awareness}), 
which leverages dynamic point clustering and structured loss objectives to guide self-supervised learning. For SeFlow, we adapt SSF~\cite{khoche2025ssf}, a fully supervised long-range scene flow method, as backbone. Finally, we include fully supervised SSF and DeFlow~\cite{zhang2024deflow} as upper-bound references to contextualize the performance of self-supervised and pseudo-labeling approaches.

\subsection{Implementation Details} \label{subsec:implementation_details}
Our method operates on fused multi-radar and multi-\gls{lidar} point cloud data, which are synchronized and projected into a shared ego-vehicle frame using the provided extrinsic calibrations.
For dynamic radar point classification, we set the Doppler threshold to $\delta_{\text{dyn}} = 0.05\,\si{m/s}$. The CCL clustering is implemented using spatial and Doppler thresholds of \(\delta_{\text{spatial}}=3.0\,\si{m}\) and \(\delta_{\text{velocity}}=1.5\,\si{m/s}\), as defined in Eq.~\ref{eq:ccl}.
For \gls{lidar} clustering, 
% the intensity threshold for noise removal is set at $0.008$. 
$\delta_{\text{intensity}}$ and $\delta_{\text{neighbor}}$ are set at $0.008$ and $0.5\,\si{m}$. 
The range-adaptive radar-to-\gls{lidar} association uses a linear interpolation between $\delta^{\text{min}}_{\text{adaptive}} = 0.1\,\si{m}$ and $\delta^{\text{max}}_{\text{adaptive}} = 5.0\,\si{m}$. All experiments are conducted on a desktop with a single NVIDIA RTX 3090 GPU. The average runtime per frame for direct scene flow estimation is $2291\pm465\,\si{ms}$, 
while for the feedforward model trained using our pseudo labels is $25\pm12\,\si{ms}$.
The peak memory usage remains below $3.5\,\si{GB}$.
Our code is implemented in PyTorch and will be made publicly available upon publication.

%% file: tables/table_1.tex
\begin{table*}[h]
\centering
\def\arraystretch{1.2}
\caption{Comparison on the TruckScenes \texttt{val} set. The first two rows are fully supervised methods, and the rest are self-supervised methods. Our method achieves state-of-the-art performance compared to self-supervised baselines. Best results are in bold, second-best are underlined.}
\label{tab:table_1}
\begin{tabular}{cc|cc|cc|c|cc}
% \begin{tabular}{ccccccccc}
\toprule
\multirow{2}{*}{Methods} & \multirow{2}{*}{Labels} & \multicolumn{2}{c|}{Range-Wise Dynamic EPE} & \multicolumn{2}{c|}{Range-wise Dynamic IoU} & Three-way EPE & Memory & Runtime \\ & & 0-35~(\si{m}) ↓ & 35+~(\si{m}) ↓ & 0-35~($\%$) ↑ & 35+~($\%$) ↑ & Mean~(\si{m}) ↓ & (MB) ↓ & (\si{ms}) ↓ \\ 
\hline
DeFlow~\cite{zhang2024deflow} & \checkmark & 0.4130 & 0.4571 & 70.19 & 58.58 & 0.1185 & 16058 & 11 $\pm$ 25 \\
SSF~\cite{khoche2025ssf} & \checkmark & 0.3966 & 0.4158 & 72.28 & 60.79 & 0.1126 & 2132 & 25 $\pm$ 12 \\
\hline
NSFP~\cite{li2021neural} & & \underline{0.6152} & 1.3442 & 39.50 & 18.48 & 0.3545 & 1894 & 28058 $\pm$ 16561 \\
FastNSF~\cite{li2023fast} & & \textbf{0.6019} & 1.0089 & 39.63 & 18.05 & \underline{0.3334} & 19268 & 2560 $\pm$ 1070 \\
ICP-Flow~\cite{lin2024icp} & & 0.8836 & 1.0994 & 35.33 & 19.96 & 0.3434 & 7193 & 5500 $\pm$ 1509 \\
SeFlow~\cite{zhang2024seflow} & & 1.0443 & \underline{0.8829} & \underline{43.30} & \underline{28.28} & 0.3369 & 2132 & 25 $\pm$ 12 \\
DoGFlow (Ours) & & 0.6892 & \textbf{0.7013} & \textbf{58.25} & \textbf{57.16} & \textbf{0.2718} & 3232 & 2291 $\pm$ 465 \\
\bottomrule
\end{tabular}
\vspace{-1em}
\end{table*}

%% file: sections/expt_results.tex
\section{Experimental Results}
\label{sec:expt_results}
\subsection{Quantitative Results}
\Cref{tab:table_1} shows the comparison of DoGFlow against state-of-the-art self-supervised scene flow methods. DoGFlow achieves the lowest mean three-way \gls{epe} among self-supervised methods, showing an 18.4\% improvement over FastNSF, the next best method. On the range-wise dynamic \gls{epe}, DoGFlow performs sligtly worse than NSFP and FastNSF at close-range (0-35\,\si{m}), but outperforms all self-supervised methods at far-range (35+\,\si{m}). 
We attribute the lower performance at close-range to the low Doppler resolution of 4D radar, which makes it difficult to resolve motion of slow-moving objects, such as pedestrians. Secondly, typical autonomous platforms can exhibit blindspots despite using multiple sensors. In TruckScenes, we observed such blindspots in sensing Doppler signals near the sides of the vehicle, limiting the ability to detect lateral motion from nearby pedestrians or vehicles. 
% The prevalence of these slow-moving objects in the platform's blindspot further biases the error distribution toward radar’s weak spots. 
\input{figures/range_wise_plot}

DoGFlow also yields significantly higher range-wise dynamic IoU, with a 34.5\% and 102.1\% improvement at close-range and far-range respectively compared to SeFlow, the second-best method. This improvement reflects the strength of 4D radar in both detecting motion and identifying static points, thereby reducing false positives in motion estimation. The most striking observation is the consistent performance across close and far-range. A detailed breakdown across range bins is shown in~\Cref{fig:range_wise_plot}. Ideally, scene flow estimation should degrade minimally with range, a trend followed closely by the supervised baseline, SSF. While NSFP and FastNSF perform well at close range (0–35 m), their performance drops sharply at longer distances. In contrast, DoGFlow exhibits smoother degradation, maintaining competitive accuracy up to and beyond 100 m. On the dynamic IoU metric, DoGFlow approaches the supervised SSF performance at long range, indicating that cross-modal label transfer from radar provides strong generalization under range-limited LiDAR conditions. In comparison, optimization-based methods that rely on Chamfer distance suffer from occlusion or missing information at long
range, leading to unstable motion estimates.
In terms of memory and runtime, DoGFlow offers a favorable trade-off between scalability and accuracy among label-free methods.

% In terms of runtime, we focus on comparisons with training-free methods, where inference speed directly affects both online scene flow estimation and offline pseudo-label generation. Although neural network-based approaches such as DeFlow, SSF, and SeFlow offer low latency at runtime, they require extensive supervised training on large-scale datasets. Among the training-free baselines, DoGFlow achieves the fastest inference speed, improving upon FastNSF by approximately 10\%. The dominant computational components in DoGFlow are noise-resistant LiDAR clustering and least-squares velocity estimation, accounting for \SI{1900}{ms} and \SI{256}{ms}, respectively. Both modules are amenable to GPU acceleration for real-time deployment. In terms of memory usage, DoGFlow scales linearly with the number of LiDAR points and quadratically with the number of dynamic radar points, due to the pairwise association and clustering stages. In practice, we empirically observe moderate memory usage—slightly higher than SSF and SeFlow, but substantially lower than optimization-based methods such as FastNSF and ICP-Flow.
% % Despite being moderately more memory-intensive than SSF and SeFlow, DoGFlow remains significantly lighter than optimization-based approaches such as FastNSF and ICP-Flow. 
% Overall, DoGFlow offers a favorable trade-off between scalability and accuracy among label-free methods.

% \input{tables/table_runtime_breakdown}

\subsection{Pseudo-Labeling Evaluation} \label{subsec:pseudo_labeling_evaluation} 
\Cref{tab:table_2} compares the SSF backbone trained on pseudo-labels generated by different training-free methods, including ICP-Flow, FastNSF, and our method DoGFlow. The fully supervised SSF model is included for reference as an upper-bound. Each method is used to generate pseudo-labels for the TruckScenes training set, which are then used to train SSF using the same protocol.
DoGFlow achieves the best performance across both near (0–35~\si{m}) and far (35+~\si{m}) ranges in terms of dynamic \gls{epe} and dynamic IoU. This demonstrates the effectiveness of radar-based label transfer in generating high-quality motion labels that generalize well across spatial scales. Compared to FastNSF and ICP-Flow, DoGFlow leads to a 36.2\% and 39.7\% improvement in far-range dynamic \gls{epe}, respectively.

\input{tables/table_2}
\input{tables/table_3}
\input{tables/experiment_4}

%% In terms of label generation efficiency, DoGFlow completes pseudo-labeling in under 65 hours, faster than FastNSF (72.5 hours) and significantly faster than ICP-Flow (156 hours). This highlights DoGFlow’s practicality for scalable scene flow annotation.
%More broadly, we observe that training an off-the-shelf backbone on noisy pseudo-labels significantly outperforms the self-supervised methods used to generate those labels. This follows a well-established trend in self-supervised learning: when pseudo-label noise is structured and informed by physical priors, learning systems can generalize effectively.

% \textbf{Dynamic Classification Evaluation:} 
We also investigate the utility of our radar-based dynamic classification by substituting it into the SeFlow pipeline, replacing DUFOMap~\cite{duberg2024dufomap}. As shown in~\Cref{tab:table_3}, this modification leads to consistent improvements in both near- and far-range dynamic \gls{epe} and IoU.
DUFOMap relies on visibility-based ray casting and map reasoning, which may struggle in adverse weather or near-field occlusion scenarios due to degraded \gls{lidar} data. In contrast, our radar-based method uses Doppler measurements to directly detect dynamic motion, offering robustness in scenarios where \gls{lidar} returns are unreliable or sparse.

% \textbf{Labeling Efficiency Evaluation}: Previously, researchers~\cite{vedder2023zeroflow} have noted that self-supervised training could surpass fully-supervised training in performance, given that the unlabeled training set is diverse and large enough in scale. Within our experimental setup, as the the unlabeled data is correlated with labeled data, we do not expect this trend to hold. Instead, we evaluate what percentage of labels are needed for finetuning beyond self-supervised learning, to achieve the same performance as fully-supervised learning.

\subsection{Label Efficiency Evaluation} \label{subsec:label_efficiency}

To assess the label efficiency enabled by DoGFlow, we investigate how much ground truth (GT) is required to reach strong scene flow performance. We compare two settings: (\textit{i}) training SSF from scratch using a varying percentage of GT annotations, and (\textit{ii}) pretraining SSF on DoGFlow pseudo-labels followed by finetuning with the same percentage of GT. The results, shown in~\Cref{fig:epe_vs_gt_label_efficiency}, report range-wise dynamic mean \gls{epe} as a function of GT fraction.

DoGFlow pseudo-labels consistently improve performance across all supervision levels. With only 10\% ground truth supervision, the DoGFlow-pretrained model achieves a mean dynamic EPE of 0.4541, nearly matching the fully supervised SSF model (0.4119). This corresponds to over 90\% of the fully supervised performance while requiring only 10\% annotations. The zero-shot (0\% GT) performance of the DoGFlow-pretrained SSF (0.5826) is competitive with SSF trained using 5\% ground truth (0.5451), highlighting its potential for zero-shot deployment when ground truth is unavailable.

These results highlight that high-quality, physics-informed pseudo labels can significantly reduce the burden of manual annotation. In practical deployments, where collecting GT across diverse cities, weather, and edge-case scenarios is prohibitively expensive, our framework offers a scalable alternative, enabling effective adaptation using only a small fraction of labeled data.

\input{figures/epe_vs_gt_label_efficiency}

\subsection{Robustness in Adverse Weather Conditions} \label{subsec:weather_robustness}

\Cref{tab:weather_comparison} reports the mean range-wise dynamic \gls{epe} and IoU of various methods under different weather conditions. DoGFlow provides the lowest dynamic \gls{epe} in Clear, Overcast, and Snow scenes.
% DoGFlow outperforms all self-supervised baselines on dynamic IoU across every weather condition and provides the lowest dynamic \gls{epe} in Clear, Overcast, and Snow scenes. 
% These results highlight the reliability of radar-based supervision under adverse conditions where LiDAR-only approaches often struggle.
In particular, self-supervised methods that rely on Chamfer-based loss functions, such as NSFP, FastNSF, and ICP-Flow, show significant performance degradation in adverse weather. 
% This is expected, as fog, snow, and rain interfere with the \gls{lidar} sensing process by scattering and absorbing laser beams. The result is missing returns, spurious measurements, and corrupted local geometry—conditions that violate the assumptions behind nearest-neighbor correspondence-based losses, ultimately leading to poor scene flow estimates.
DoGFlow’s robustness in these scenarios stems from its use of radar-derived labels and Doppler-based dynamic classification. Unlike \gls{lidar}, millimeter-wave radar is minimally affected by adverse weather and continues to provide reliable velocity measurements. This makes DoGFlow less sensitive to the geometric degradation that plagues \gls{lidar}-only methods.

Interestingly, SeFlow achieves slightly better \gls{epe} than DoGFlow in rain and fog. This can be attributed to DUFOMap's time-integrated ray-casting strategy. By aggregating observations over multiple frames and relying on voxel-based occupancy violations, DUFOMap is able to filter out short-lived noise from atmospheric particles and produce stable dynamic masks even in noisy environments. While DoGFlow’s Doppler thresholding is simpler and faster, it is more susceptible to multipath effects and weak signals, particularly in slow-moving objects or cluttered environments.

Despite this, DoGFlow significantly outperforms SeFlow and all other baselines in dynamic IoU across all weather conditions. This suggests that radar-based label transfer
% motion supervision 
excels at distinguishing moving from static entities, even when exact motion vectors may be slightly noisy. It also underscores the importance of our noise-resistant LiDAR clustering and label propagation stages, which help maintain performance under sensor corruption.

%% file: figures/range_wise_plot.tex
\begin{figure}[ht]
\centering
\includegraphics[trim=0 0 0 0, clip, width=0.99\linewidth]{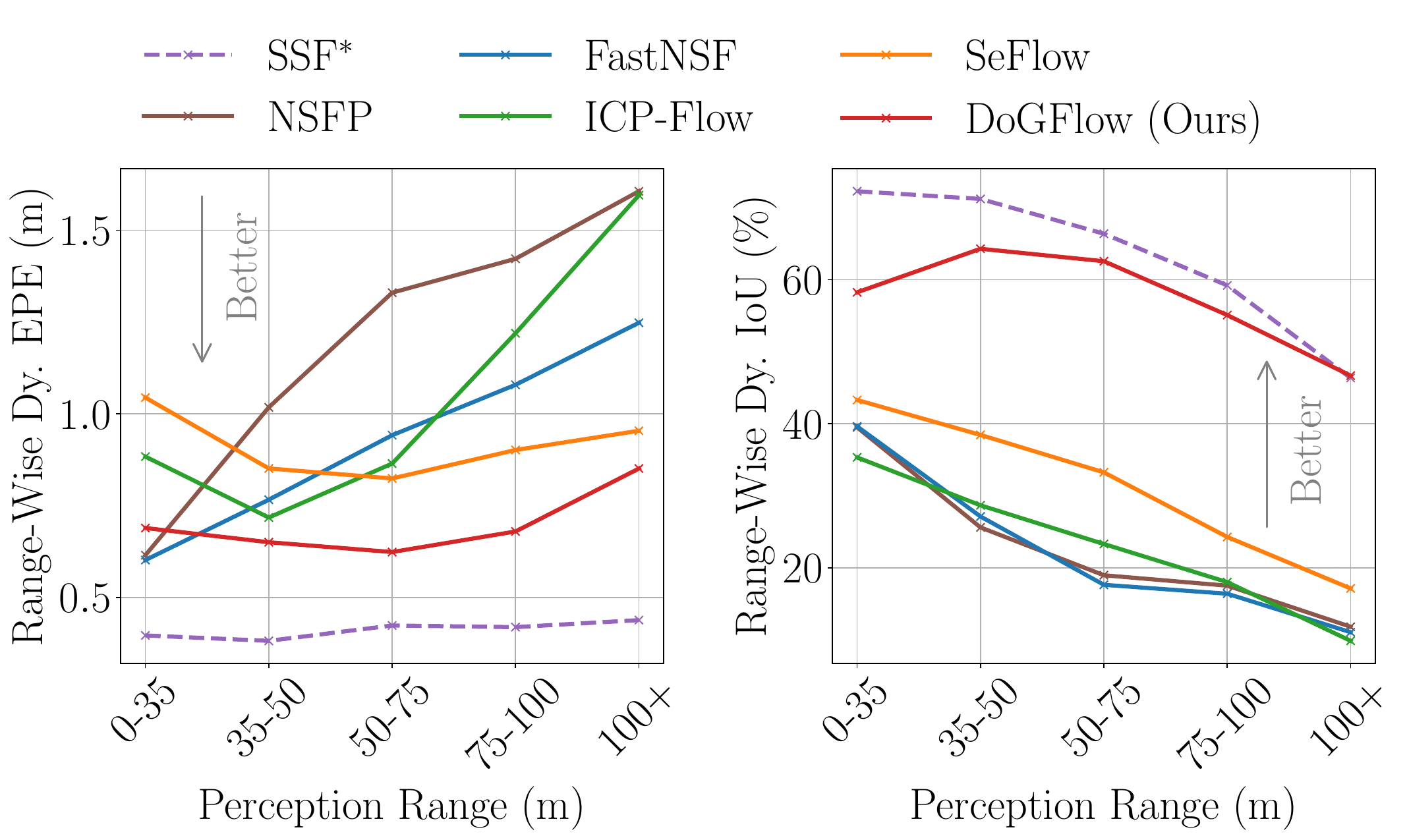}
\vspace{-1em}
\caption{Range-wise breakdown of scene flow performance. Left: Dynamic \gls{epe} and Right: Dynamic IoU across perception range bins. Our method (DogFlow, red) exhibits significantly lower degradation with increasing range compared to prior methods. SSF$^*$: Fully supervised. }
%, demonstrating robustness in long-range scenarios.}
\label{fig:range_wise_plot}
\vspace{-1em}
\end{figure}

%% file: tables/table_2.tex
\begin{table}[h]
\centering
\def\arraystretch{1.2}
\caption{Comparison of the SSF~\cite{khoche2025ssf} backbone trained using pseudo-labels generated by various training-free methods. DoGFlow provides the best performance in both near- and far-range metrics while also requiring the least time to generate labels.}
\label{tab:table_2}
\resizebox{\columnwidth}{!}{  % Resize to fit within one column
\begin{tabular}{c|cc|cc}
\toprule
\multirow{2}{*}{Method} & \multicolumn{2}{c|}{Range-Wise Dy. EPE} & \multicolumn{2}{c}{Range-wise Dy. IoU} \\
& 0-35~(\si{m}) ↓ & 35+~(\si{m}) ↓ & 0-35~($\%$) ↑ & 35+~($\%$) ↑ \\
\hline
\rowcolor[rgb]{0.91,0.91,0.91} Ground-truth  & 0.3966 & 0.4158 & 72.28 & 60.79 \\
FastNSF~\cite{li2023fast} & 0.5569 & 0.9283 & 43.48 & 35.78 \\
ICP-Flow~\cite{lin2024icp} & 0.8125 & 0.9827 & 38.46 & 37.58 \\
DoGFlow (Ours) & \textbf{0.5451} & \textbf{0.5920} & \textbf{65.86} & \textbf{49.03} \\
\bottomrule
\end{tabular}
}
\vspace{-1em}
\end{table}

% \begin{table*}[h]
% \centering
% \def\arraystretch{1.2}
% \caption{Comparison of the SSF backbone trained using pseudo-labels generated by various training-free methods. RSSF provides the best performance in both near- and far-range metrics while also requiring the least time to generate labels.}
% \label{tab:table_2}
% \begin{tabular}{c|c|c|cc|cc}
% \toprule
% \multirow{2}{*}{Backbone} & \multirow{2}{*}{Method} & Pseudo-labeling & \multicolumn{2}{c|}{Range-Wise Dynamic EPE} & \multicolumn{2}{c}{Range-wise Dynamic IoU} \\
% & & time (hours) & 0-35~(\si{m}) ↓ & 35+~(\si{m}) ↓ & 0-35~($\%$) ↑ & 35+~($\%$) ↑ \\
% \hline
% \multirow{3}{*}{SSF~\cite{khoche2025ssf}} & \cellcolor[rgb]{0.91,0.91,0.91} Ground-truth & \cellcolor[rgb]{0.91,0.91,0.91} -    & \cellcolor[rgb]{0.91,0.91,0.91} 0.3966 & \cellcolor[rgb]{0.91,0.91,0.91} 0.4158 & \cellcolor[rgb]{0.91,0.91,0.91} 72.28 & \cellcolor[rgb]{0.91,0.91,0.91} 60.79 \\
% & FastNSF~\cite{li2023fast} & 72.46 & 0.5569 & 0.9283 & 43.48 & 35.78 \\
% & ICP-Flow~\cite{lin2024icp}    & 155.68 & 0.8125 & 0.9827 & 38.46 & 37.58 \\
% & RSSF (Ours)  & 64.82  & \textbf{0.5451} & \textbf{0.5920} & \textbf{65.86} & \textbf{49.03} \\
% \bottomrule
% \end{tabular}
% \end{table*}

%% file: tables/table_3.tex
\newcommand{\blueup}[1]{$_{\color{TableBlue}\uparrow #1}$}
\newcommand{\bluedown}[1]{$_{\color{TableBlue}\downarrow #1}$}
\begin{table}[h]
\centering
\def\arraystretch{1.2}
\caption{Evaluation of radar-based dynamic classification as a replacement for DUFOMap within the SeFlow framework. Best results are in bold.}
% Using radar improves scene flow accuracy and dynamic IoU across all ranges.}
\label{tab:table_3}
\resizebox{\columnwidth}{!}{  % Resize to fit within one column
\begin{tabular}{c|cc|cc}
\toprule
\multirow{2}{*}{Methods} & \multicolumn{2}{c|}{Range-Wise Dy. EPE} & \multicolumn{2}{c}{Range-wise Dy. IoU} \\ & 0-35~(\si{m}) ↓ & 35+~(\si{m}) ↓ & 0-35~($\%$) ↑ & 35+~($\%$) ↑ \\ 
\hline
SeFlow~\cite{zhang2024seflow} & \multicolumn{0}{l}{1.0443} & \multicolumn{0}{l}{0.8829} & \multicolumn{0}{l}{43.30} & \multicolumn{0}{l}{28.28} \\
SeFlow w. Radar & \textbf{0.8048} \bluedown{23\%} & \textbf{0.7193} \bluedown{18\%} & \textbf{48.87} \blueup{13\%} & \textbf{37.44} \blueup{32\%} \\
\bottomrule
\end{tabular}
}
\vspace{-1em}
\end{table}

%% file: tables/experiment_4.tex
\begin{table*}[h]
\centering
\def\arraystretch{1.2}
\caption{Scene Flow Performance Under Adverse Weather Conditions
% .Comparison of range-wise dynamic \gls{epe} (↓) and dynamic IoU (↑) across different weather types 
on the TruckScenes \texttt{val} set. 
% RSSF consistently achieves the highest IoU across all weather conditions and lowest EPE in clear, overcast, and snow scenes. 
Best results are in bold, second-best are underlined. SSF is \colorbox{gray!40}{supervised} method.}
\label{tab:weather_comparison}
\begin{tabular}{c|c|c|c|c|c}
% \toprule
\hline
% \multirow{2}{*}{Methods} & \multicolumn{6}{c}{Range-wise Dynamic Mean: \gls{epe}~(\si{m}) ↓ / IoU~(\%) ↑} \\ & NSFP & FastNSF & ICP-Flow & RSSF (Ours) \\
\multirow{2}{*}{Methods} & \multicolumn{5}{c}{Mean Range-Wise Dynamic:~\gls{epe}~(\si{m}) ↓ / IoU~($\%$) ↑} \\ &  Clear & Overcast & Rain & Snow & Fog \\
\hline
\rowcolor[rgb]{0.91,0.91,0.91} SSF~\cite{khoche2025ssf} & 0.4468 / 63.64 & 0.3628 / 65.00 & 0.4531 / 46.92 & 0.5624 / 65.43 & 0.2901 / 48.04 \\
NSFP~\cite{li2021neural} & 1.2883 / 21.79 & 1.0872 / 26.45 & 1.0531 / 14.85 & 1.6778 / 14.07 & 0.5503 / 12.42 \\
FastNSF~\cite{li2023fast} & 0.9699 / 21.65 & 0.8572 / 25.92 & 0.9238 / 16.09 & \underline{1.0739} / 14.16 & 0.7888 / 8.74 \\
ICP-Flow~\cite{lin2024icp} & 1.1211 / 22.75 & 0.9345 / 25.21 & 1.0938 / 17.19 & 1.7282 / 13.13 & 0.5365 / 15.11 \\
SeFlow~\cite{zhang2024seflow} & \underline{0.9683} / \underline{31.36} & \underline{0.8294} / \underline{33.46} & \textbf{0.8038} / \underline{18.82} & 1.6786 / \underline{21.81} & \textbf{0.4208} / \underline{23.82} \\
\textbf{DoGFlow (Ours)} & \textbf{0.7164} / \textbf{58.53} & \textbf{0.6423} / \textbf{58.92} & \underline{0.8108} / \textbf{39.30} & \textbf{0.9004} / \textbf{67.96} & \underline{0.5183} / \textbf{45.36} \\
% \bottomrule
\hline
\end{tabular}
\vspace{-1em}
\end{table*}

%% file: figures/epe_vs_gt_label_efficiency.tex
\begin{figure}[ht]
\centering
\includegraphics[trim=0 0 0 0, clip, width=0.99\linewidth]{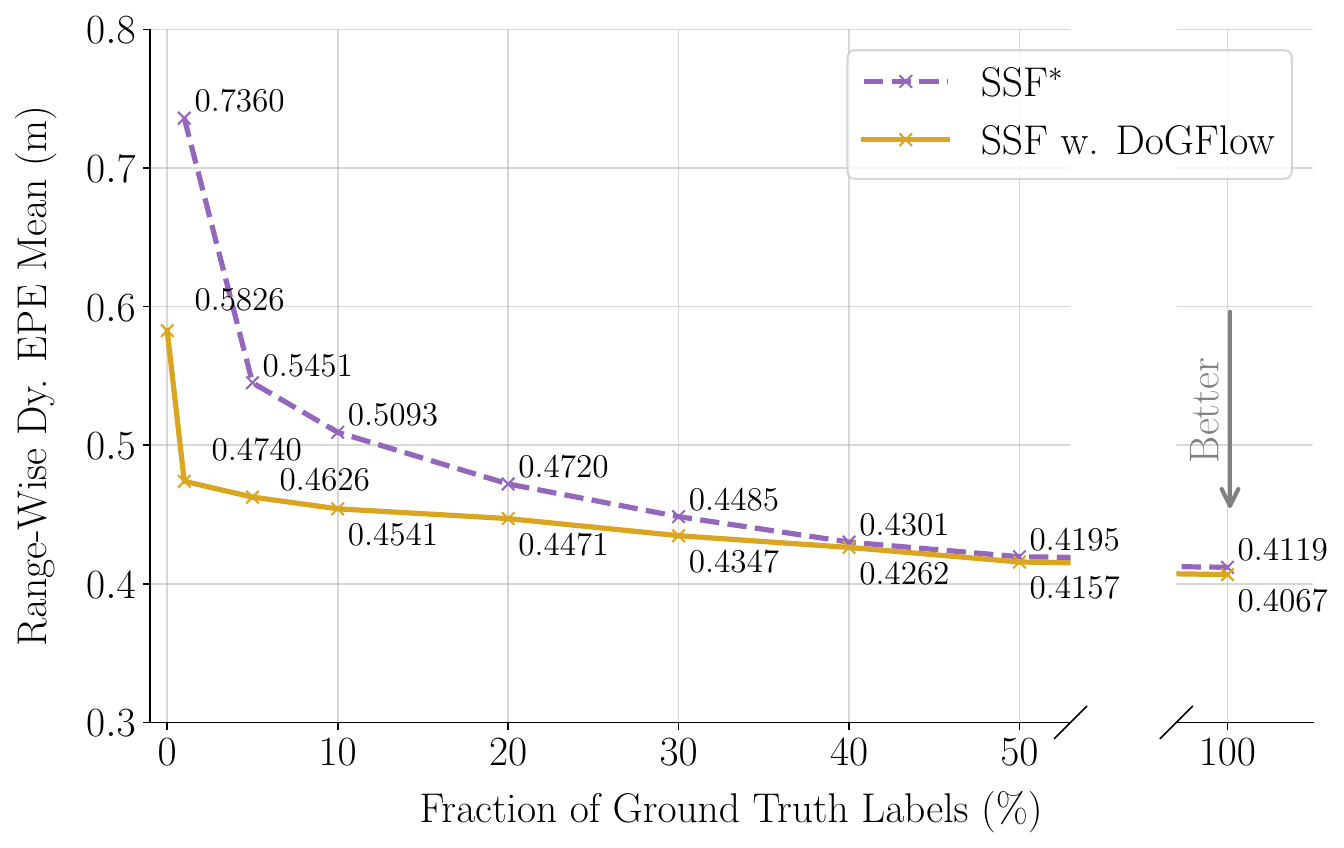}
\vspace{-2em}
\caption{%
Label efficiency evaluation comparing SSF trained from scratch (purple) versus SSF pretrained on DoGFlow pseudo labels and finetuned with limited ground truth (gold). DoGFlow pretraining significantly boosts performance in low-label regimes. With only 10\% ground truth, the pretrained model achieves over 90\% of fully supervised performance, reducing annotation cost while maintaining accuracy.
}
\label{fig:epe_vs_gt_label_efficiency}
\vspace{-1em}
\end{figure}

%% file: sections/conclusions.tex
\section{Conclusion} \label{sec:conclusions}
We presented DoGFlow, a cross-modal label transfer framework for \gls{lidar} scene flow estimation that leverages Doppler-based velocity cues and clustering-based rigidity priors to generate accurate scene flow estimation. Our method combines the complementary strengths of 4D radar and \gls{lidar} to produce robust motion estimates across both short- and long-range scenes, even under adverse weather conditions. Through extensive experiments on the MAN TruckScenes dataset, we demonstrate that DoGFlow sets a new benchmark among self-supervised methods, while also enabling effective pseudo-labeling for downstream training of real-time scene flow backbones. Furthermore, our label-efficiency analysis shows that DoGFlow substantially reduces reliance on ground truth annotations, enabling scalable deployment in diverse and unlabeled environments.

The method assumes accurate and static extrinsic calibration between radar and LiDAR. Sensor drift or slight misalignment could introduce systematic velocity errors. Additional limitations include degraded close-range motion accuracy in sensor blindspots or regions with weak Doppler returns. Future work could explore integrating complementary priors from existing self-supervised methods to enhance performance in these challenging areas. 
% While we discuss limitations such as degraded close-range motion accuracy in some sensor blindspots or weak Doppler conditions, our experiments show that these are more than offset by the improved generalization and robustness in the long tail of real-world scenes. 
% Future work may explore 
% % tighter radar-\gls{lidar} temporal synchronization,
% multi-frame aggregation or confidence-weighted fusion strategies to further enhance pseudo-label quality and runtime applicability.

%% file: main.bbl
\begin{thebibliography}{10}
\providecommand{\url}[1]{#1}
\csname url@samestyle\endcsname
\providecommand{\newblock}{\relax}
\providecommand{\bibinfo}[2]{#2}
\providecommand{\BIBentrySTDinterwordspacing}{\spaceskip=0pt\relax}
\providecommand{\BIBentryALTinterwordstretchfactor}{4}
\providecommand{\BIBentryALTinterwordspacing}{\spaceskip=\fontdimen2\font plus
\BIBentryALTinterwordstretchfactor\fontdimen3\font minus \fontdimen4\font\relax}
\providecommand{\BIBforeignlanguage}[2]{{%
\expandafter\ifx\csname l@#1\endcsname\relax
\typeout{** WARNING: IEEEtran.bst: No hyphenation pattern has been}%
\typeout{** loaded for the language `#1'. Using the pattern for}%
\typeout{** the default language instead.}%
\else
\language=\csname l@#1\endcsname
\fi
#2}}
\providecommand{\BIBdecl}{\relax}
\BIBdecl

\bibitem{lentsch2024union}
T.~Lentsch, H.~Caesar, and D.~Gavrila, ``Union: Unsupervised 3d object detection using object appearance-based pseudo-classes,'' \emph{Advances in Neural Information Processing Systems}, vol.~37, pp. 22\,028--22\,046, 2024.

\bibitem{pan2024ratrack}
Z.~Pan, F.~Ding, H.~Zhong, and C.~X. Lu, ``Ratrack: moving object detection and tracking with 4d radar point cloud,'' in \emph{2024 IEEE International Conference on Robotics and Automation (ICRA)}.\hskip 1em plus 0.5em minus 0.4em\relax IEEE, 2024, pp. 4480--4487.

\bibitem{chen2025semanticflow}
Y.~Chen, M.~Zhang, Q.~Hao, and G.~Zhou, ``Semanticflow: A self-supervised framework for joint scene flow prediction and instance segmentation in dynamic environments,'' \emph{arXiv preprint arXiv:2503.14837}, 2025.

\bibitem{murhij2024ofmpnet}
Y.~Murhij and D.~Yudin, ``Ofmpnet: Deep end-to-end model for occupancy and flow prediction in urban environment,'' \emph{Neurocomputing}, vol. 586, p. 127649, 2024.

\bibitem{jund2021scalable}
P.~Jund, C.~Sweeney \emph{et~al.}, ``Scalable scene flow from point clouds in the real world,'' \emph{IEEE Robotics and Automation Letters}, vol.~7, no.~2, pp. 1589--1596, 2021.

\bibitem{zhang2024deflow}
Q.~Zhang, Y.~Yang \emph{et~al.}, ``{DeFlow}: Decoder of scene flow network in autonomous driving,'' in \emph{2024 IEEE International Conference on Robotics and Automation (ICRA)}, 2024, pp. 2105--2111.

\bibitem{khoche2025ssf}
A.~Khoche, Q.~Zhang \emph{et~al.}, ``Ssf: Sparse long-range scene flow for autonomous driving,'' \emph{arXiv preprint arXiv:2501.17821}, 2025.

\bibitem{kim2025flow4d}
J.~Kim, J.~Woo \emph{et~al.}, ``Flow4d: Leveraging 4d voxel network for lidar scene flow estimation,'' \emph{IEEE Robotics and Automation Letters}, 2025.

\bibitem{khoche2024towards}
A.~Khoche, L.~P. S{\'a}nchez \emph{et~al.}, ``Towards long-range 3d object detection for autonomous vehicles,'' in \emph{2024 IEEE Intelligent Vehicles Symposium (IV)}.\hskip 1em plus 0.5em minus 0.4em\relax IEEE, 2024, pp. 2206--2212.

\bibitem{vedder2023zeroflow}
K.~Vedder, N.~Peri \emph{et~al.}, ``Zeroflow: Scalable scene flow via distillation,'' \emph{arXiv preprint arXiv:2305.10424}, 2023.

\bibitem{vedder2024neural}
------, ``Neural eulerian scene flow fields,'' in \emph{The Thirteenth International Conference on Learning Representations}, 2025.

\bibitem{li2021neural}
X.~Li, J.~Kaesemodel~Pontes, and S.~Lucey, ``Neural scene flow prior,'' \emph{Advances in Neural Information Processing Systems}, vol.~34, pp. 7838--7851, 2021.

\bibitem{zhang2024seflow}
Q.~Zhang, Y.~Yang \emph{et~al.}, ``{SeFlow}: A self-supervised scene flow method in autonomous driving,'' in \emph{European Conference on Computer Vision (ECCV)}.\hskip 1em plus 0.5em minus 0.4em\relax Springer, 2024, p. 353–369.

\bibitem{mittal2020just}
H.~Mittal, B.~Okorn, and D.~Held, ``Just go with the flow: Self-supervised scene flow estimation,'' in \emph{Proceedings of the IEEE/CVF conference on computer vision and pattern recognition}, 2020, pp. 11\,177--11\,185.

\bibitem{zhang2025himo}
Q.~Zhang, A.~Khoche \emph{et~al.}, ``{HiMo}: High-speed objects motion compensation in point cloud,'' \emph{arXiv preprint arXiv:2503.00803}, 2025.

\bibitem{luo2021self}
C.~Luo, X.~Yang, and A.~Yuille, ``Self-supervised pillar motion learning for autonomous driving,'' in \emph{Proceedings of the IEEE/CVF Conference on Computer Vision and Pattern Recognition}, 2021, pp. 3183--3192.

\bibitem{jiang20223}
C.~Jiang, G.~Wang, Y.~Miao, and H.~Wang, ``3-d scene flow estimation on pseudo-lidar: Bridging the gap on estimating point motion,'' \emph{IEEE Transactions on Industrial Informatics}, vol.~19, no.~6, pp. 7346--7354, 2022.

\bibitem{hoffmann2025floxels}
D.~T. Hoffmann, S.~H. Raza \emph{et~al.}, ``Floxels: Fast unsupervised voxel based scene flow estimation,'' \emph{arXiv preprint arXiv:2503.04718}, 2025.

\bibitem{Dreissig2023SurveyOL}
\BIBentryALTinterwordspacing
M.~Dreissig, D.~Scheuble, F.~Piewak, and J.~Boedecker, ``Survey on lidar perception in adverse weather conditions,'' \emph{2023 IEEE Intelligent Vehicles Symposium (IV)}, pp. 1--8, 2023. [Online]. Available: \url{https://api.semanticscholar.org/CorpusID:258108386}
\BIBentrySTDinterwordspacing

\bibitem{wang2022scalable}
W.~Wang, X.~You \emph{et~al.}, ``A scalable and accurate de-snowing algorithm for lidar point clouds in winter,'' \emph{Remote Sensing}, vol.~14, no.~6, p. 1468, 2022.

\bibitem{yu2022lisnownet}
M.-Y. Yu, R.~Vasudevan, and M.~Johnson-Roberson, ``Lisnownet: Real-time snow removal for lidar point clouds,'' in \emph{2022 IEEE/RSJ International Conference on Intelligent Robots and Systems (IROS)}.\hskip 1em plus 0.5em minus 0.4em\relax IEEE, 2022, pp. 6820--6826.

\bibitem{ding2022self}
F.~Ding, Z.~Pan \emph{et~al.}, ``Self-supervised scene flow estimation with 4-d automotive radar,'' \emph{IEEE Robotics and Automation Letters}, vol.~7, no.~3, pp. 8233--8240, 2022.

\bibitem{caesar2020nuscenes}
H.~Caesar, V.~Bankiti \emph{et~al.}, ``nuscenes: A multimodal dataset for autonomous driving,'' in \emph{Proceedings of the IEEE/CVF conference on computer vision and pattern recognition}, 2020, pp. 11\,621--11\,631.

\bibitem{wilson2argoverse}
B.~Wilson, W.~Qi \emph{et~al.}, ``Argoverse 2: Next generation datasets for self-driving perception and forecasting,'' in \emph{Thirty-fifth Conference on Neural Information Processing Systems Datasets and Benchmarks Track}.

\bibitem{fent2024man}
F.~Fent, F.~Kuttenreich \emph{et~al.}, ``Man truckscenes: A multimodal dataset for autonomous trucking in diverse conditions,'' \emph{Advances in Neural Information Processing Systems}, vol.~37, pp. 62\,062--62\,082, 2024.

\bibitem{lin2024icp}
Y.~Lin and H.~Caesar, ``Icp-flow: Lidar scene flow estimation with icp,'' in \emph{Proceedings of the IEEE/CVF Conference on Computer Vision and Pattern Recognition}, 2024, pp. 15\,501--15\,511.

\bibitem{li2023fast}
X.~Li, J.~Zheng \emph{et~al.}, ``Fast neural scene flow,'' in \emph{Proceedings of the IEEE/CVF International Conference on Computer Vision}, 2023, pp. 9878--9890.

\bibitem{li2022rigidflow}
R.~Li, C.~Zhang \emph{et~al.}, ``Rigidflow: Self-supervised scene flow learning on point clouds by local rigidity prior,'' in \emph{Proceedings of the IEEE/CVF Conference on Computer Vision and Pattern Recognition}, 2022, pp. 16\,959--16\,968.

\bibitem{duberg2024dufomap}
D.~Duberg, Q.~Zhang, M.~Jia, and P.~Jensfelt, ``Dufomap: Efficient dynamic awareness mapping,'' \emph{IEEE Robotics and Automation Letters}, 2024.

\bibitem{yang2023emernerf}
J.~Yang, B.~Ivanovic \emph{et~al.}, ``Emernerf: Emergent spatial-temporal scene decomposition via self-supervision,'' \emph{arXiv preprint arXiv:2311.02077}, 2023.

\bibitem{long2021full}
Y.~Long, D.~Morris \emph{et~al.}, ``Full-velocity radar returns by radar-camera fusion,'' in \emph{Proceedings of the IEEE/CVF International Conference on Computer Vision}, 2021, pp. 16\,198--16\,207.

\bibitem{ding2023hidden}
F.~Ding, A.~Palffy, D.~M. Gavrila, and C.~X. Lu, ``Hidden gems: 4d radar scene flow learning using cross-modal supervision,'' in \emph{Proceedings of the IEEE/CVF Conference on Computer Vision and Pattern Recognition}, 2023, pp. 9340--9349.

\bibitem{vizzo2023kiss}
I.~Vizzo, T.~Guadagnino \emph{et~al.}, ``Kiss-icp: In defense of point-to-point icp--simple, accurate, and robust registration if done the right way,'' \emph{IEEE Robotics and Automation Letters}, vol.~8, no.~2, pp. 1029--1036, 2023.

\bibitem{wang2021cfar}
J.~Wang, ``Cfar-based interference mitigation for fmcw automotive radar systems,'' \emph{IEEE Transactions on Intelligent Transportation Systems}, vol.~23, no.~8, pp. 12\,229--12\,238, 2021.

\bibitem{levy2023mcrb}
M.~Levy-Israel, I.~Bilik, and J.~Tabrikian, ``Mcrb on doa estimation for automotive mimo radar in the presence of multipath,'' \emph{IEEE Transactions on Aerospace and Electronic Systems}, vol.~59, no.~5, pp. 4831--4843, 2023.

\bibitem{kopp2021fast}
J.~Kopp, D.~Kellner, A.~Piroli, and K.~Dietmayer, ``Fast rule-based clutter detection in automotive radar data,'' in \emph{2021 IEEE International Intelligent Transportation Systems Conference (ITSC)}.\hskip 1em plus 0.5em minus 0.4em\relax IEEE, 2021, pp. 3010--3017.

\bibitem{mcinnes2017hdbscan}
L.~McInnes, J.~Healy, S.~Astels \emph{et~al.}, ``hdbscan: Hierarchical density based clustering.'' \emph{J. Open Source Softw.}, vol.~2, no.~11, p. 205, 2017.

\bibitem{fan2022fully}
L.~Fan, F.~Wang, N.~Wang, and Z.-X. Zhang, ``Fully sparse 3d object detection,'' \emph{Advances in Neural Information Processing Systems}, vol.~35, pp. 351--363, 2022.

\bibitem{branch1999subspace}
M.~A. Branch, T.~F. Coleman, and Y.~Li, ``A subspace, interior, and conjugate gradient method for large-scale bound-constrained minimization problems,'' \emph{SIAM Journal on Scientific Computing}, vol.~21, no.~1, pp. 1--23, 1999.

\bibitem{steinke2023groundgrid}
N.~Steinke, D.~Goehring, and R.~Rojas, ``Groundgrid: Lidar point cloud ground segmentation and terrain estimation,'' \emph{IEEE Robotics and Automation Letters}, vol.~9, no.~1, pp. 420--426, 2023.

\bibitem{everingham2015pascal}
M.~Everingham, S.~A. Eslami \emph{et~al.}, ``The pascal visual object classes challenge: A retrospective,'' \emph{International journal of computer vision}, vol. 111, pp. 98--136, 2015.

\bibitem{Chodosh_2024_WACV}
N.~Chodosh, D.~Ramanan, and S.~Lucey, ``Re-evaluating lidar scene flow,'' in \emph{Proceedings of the IEEE/CVF Winter Conference on Applications of Computer Vision (WACV)}, January 2024, pp. 6005--6015.

\end{thebibliography}
